\documentclass{article}

\usepackage[numbers]{natbib}
\usepackage{multibib}
\newcites{methods}{Methods References}
\usepackage{makecell}
\usepackage[utf8]{inputenc}
\usepackage{xspace}
\usepackage{graphicx}
\usepackage{float}
\usepackage{amsfonts}
\usepackage{amsmath}
\usepackage{bbm}
\usepackage[dvipsnames,table,xcdraw]{xcolor}

\usepackage{graphicx}
\usepackage[labelfont=bf]{caption}

\usepackage[colorlinks,linktoc=all]{hyperref}
\usepackage[all]{hypcap}
\hypersetup{citecolor=MidnightBlue}
\hypersetup{linkcolor=MidnightBlue}
\hypersetup{urlcolor=MidnightBlue}

\usepackage[capitalize,nameinlink]{cleveref}
\usepackage{multirow}
\usepackage{graphicx}
\usepackage[margin=1in]{geometry}
\usepackage{booktabs}
\usepackage[T1]{fontenc}
\usepackage[acronym,smallcaps, shortcuts,nowarn,nohypertypes={acronym,notation}]{glossaries}

\usepackage{caption}
\usepackage{subcaption}
\newacronym{cnn}{cnn}{convolutional neural network}

\newacronym{icu}{icu}{intensive care unit}

\newacronym{ehr}{ehr}{electronic health record}

\newacronym{nyu}{nyu}{New York University}

\newacronym{acs}{acs}{acute coronary syndrome}

\newacronym{auroc}{auroc}{area under the receiver operator characteristic curve}

\newacronym{auc}{auc}{area under curve}

\newacronym{roc}{roc}{receiving operator characteristic}

\newacronym{rnn}{rnn}{recurrent neural network}

\newacronym{apache}{apache}{Acute Physiology and Chronic Health Evaluation}

\newacronym{icd-9}{icd-9}{International Classification of Diseases, 9th Revision}

\newacronym{icd}{icd}{International Classification of Diseases}

\newacronym{irb}{irb}{Institutional Review Boards}

\DeclareRobustCommand{\indicator}[1]{\ensuremath{\mathbbm{1}\left[#1\right]}}

\usepackage{lineno}
\date{}
\begin{document}
\title{A dynamic risk score for early prediction of cardiogenic shock using machine learning}
\author{Yuxuan Hu$^1$, Albert Lui$^2$, Mark Goldstein$^3$, Mukund Sudarshan$^3$, \\Andrea Tinsay$^4$, Cindy Tsui$^4$, Samuel Maidman$^4$, John Medamana$^4$, Neil Jethani$^{2,3}$, \\Aahlad Puli$^{3}$, Vuthy Nguy$^5$, Yindalon Aphinyanaphongs$^5$, Nicholas Kiefer$^1$, \\Nathaniel Smilowitz$^1$, James Horowitz$^1$, Tania Ahuja$^6$, Glenn I Fishman$^1$, Judith Hochman$^1$,\\ Stuart Katz$^1$, Samuel Bernard$^1$, Rajesh Ranganath$^{3,5,7}$\\
$^1$ Leon. H. Charney Division of Cardiology, NYU Langone Health \\
$^2$ NYU Grossman School of Medicine\\
$^3$ Courant Institute of Mathematics, New York University\\
$^4$ Department of Medicine, NYU Langone Health\\
$^5$ Department of Population Health, NYU Langone Health \\
$^6$ Department of Pharmacy, NYU Langone Health \\
$^7$ Center for Data Science, New York University\\
}
\date{}
\maketitle
\section*{Abstract}
Myocardial infarction and heart failure are major cardiovascular diseases that affect millions of people in the US. The morbidity and mortality are highest among patients who develop cardiogenic shock. Early recognition of cardiogenic shock is critical. Prompt implementation of treatment measures can prevent the deleterious spiral of ischemia, low blood pressure, and reduced cardiac output due to cardiogenic shock. However, early identification of cardiogenic shock has been challenging due to human providers' inability to process the enormous amount of data in the cardiac \gls{icu} and lack of an effective risk stratification tool. We developed a deep learning-based risk stratification tool, called CShock, for patients admitted into the cardiac \gls{icu} with acute decompensated heart failure and/or myocardial infarction to predict onset of cardiogenic shock. To develop and validate CShock, we annotated cardiac \gls{icu} datasets with physician adjudicated outcomes. CShock achieved an \gls{auroc} of 0.820, which substantially outperformed CardShock (\gls{auroc} 0.519), a well-established risk score for cardiogenic shock prognosis. CShock was externally validated in an independent patient cohort and achieved an \gls{auroc} of 0.800, demonstrating its generalizability in other cardiac \glspl{icu}.

\glsresetall
\section*{Introduction}
Cardiogenic shock is a life-threatening condition that is characterized by reduced cardiac output in the presence of adequate intravascular volume, resulting in tissue hypoxia. It complicates 5\% to 12\% of myocardial infarctions \cite{goldberg2009thirty,kolte2014trends,van2017contemporary,wayangankar2016temporal} and 2\% to 5\% of acute decompensated heart failure \cite{chioncel2020epidemiology,rossello2021synergistic}, the two most common etiologies for cardiogenic shock \cite{berg2019epidemiology}. Mortality for cardiogenic shock remains high (30\%-50\%) \cite{berg2019epidemiology,hochman2003cardiogenic,hochman1999early,lawler2020clinical}. Early identification of cardiogenic shock can facilitate interventions that may mitigate the consequences of prolonged end-organ insult \cite{papolos2021management,reyentovich2016management}, such as rapid employment of hemodynamic support with pharmacologic and nonpharmacologic agents, engagement of a dedicated shock team, and prompt transfer of patients at lower-acuity hospitals to a tertiary high-volume shock hub \cite{van2017contemporary}. In the setting of acute myocardial infarction, early recognition of patients at high risk for cardiogenic shock allows providers to proceed with urgent revascularization of the infarct-related coronary artery \cite{hochman1999early}. Therefore, timely recognition of patients at high risk for cardiogenic shock is crucial for improving the care of patients and their outcomes.\\

Despite the potential benefits, early identification of cardiogenic shock has been challenging. \Gls{icu} providers are presented with tremendous amounts of data generated from multiple sources including laboratory measurements, vital signs, hemodynamics, and cardiac function studies. The limited ability of human providers to process, interpret, and act upon all the data stored in the \gls{ehr} in a timely fashion can lead to poor patient outcomes. Moreover, it is difficult to provide around-the-clock monitoring and risk assessment of cardiac \gls{icu} patients by human caregivers, especially in settings with low provider-to-patient ratios.\\ 

Furthermore, there lacks an effective risk stratification tool for development of cardiogenic shock. Current risk stratification strategies such as the Society for Cardiovascular Angiography and Intervention classification schema \cite{baran2019scai,naidu2022scai} lack specific quantitative criteria, making it challenging to apply clinically. Well-established scores such as CardShock \cite{harjola2015clinical}, IABP-shock II \cite{poss2017risk} and \gls{apache} II \cite{knaus1985apache} were developed to predict outcomes such as in-hospital or 30-day mortality rather than the development of cardiogenic shock. Recent machine learning algorithms only classified patients who were initiated on inotropes and/or mechanical circulatory support as developing cardiogenic shock, but omitted those with low blood pressure and end-organ hypoperfusion \cite{chang2022early,rahman2022using}. The best positive predictive value was 11\% (with recall/sensitivity 27\%) due to the low prevalence (1\%) of cardiogenic shock patients in the study cohort  \cite{rahman2022using}, which would result in the algorithm having a high rate of false alarms, while also missing the majority of cardiogenic shock patients.\\

To improve risk stratification of cardiac critical care patients and provide early warning of cardiogenic shock for patients admitted into cardiac \gls{icu} with 1) acute decompensated heart failure and/or 2) myocardial infarction with or without ST-elevation, we develop a new dynamic risk score, called CShock, by training a machine learning model with a novel loss function designed specifically for risk scoring. We prepared a cardiac \gls{icu} dataset using MIMIC-III (Medical Information Mart for Intensive Care) database \cite{johnson2016mimic} by annotating with physician adjudicated outcomes and relevant features manually extracted from echocardiogram and catheterization reports. This dataset was then used to train CShock. We externally validated the risk model on the \gls{nyu} Langone Health cardiac \gls{icu} database that was also annotated with physician adjudicated outcomes.\\
\begin{table}[!tb]
    \centering
    \begin{tabular}{llll}
    \toprule
    \multirow{3}{*}{\parbox{4cm}{systolic blood pressure less than 90mmHg for at least 30 minutes}} & & urine output less than 0.5cc/kg/hour for at least 6 hours or,\\
    & AND & increase in serum creatinine by 1.5 fold or 0.3mg/dl from baseline or, \\
    & & serum lactate above the upper limit of normal, which is > 2mmol/L \\
    \midrule
    \multicolumn{3}{c}{OR}\\
    \midrule
    \multicolumn{3}{l}{\parbox{16cm}{pharmacologic agents and/or mechanical circulatory
    support were initiated to maintain systolic blood pressure above 90mmHg}}\\
    \bottomrule
    \end{tabular}
    \caption{%
    \textit{Structured data used to determine time of shock onset \label{tab:hypoperfusion}}
    }
    \end{table}
\section*{Results}
\subsection*{Preparation of a cardiac \gls{icu} dataset with physician adjudicated outcomes and cardiac features extracted from unstructured data} 
\begin{figure}[!ht]
    \centering
{\includegraphics[width=\textwidth]{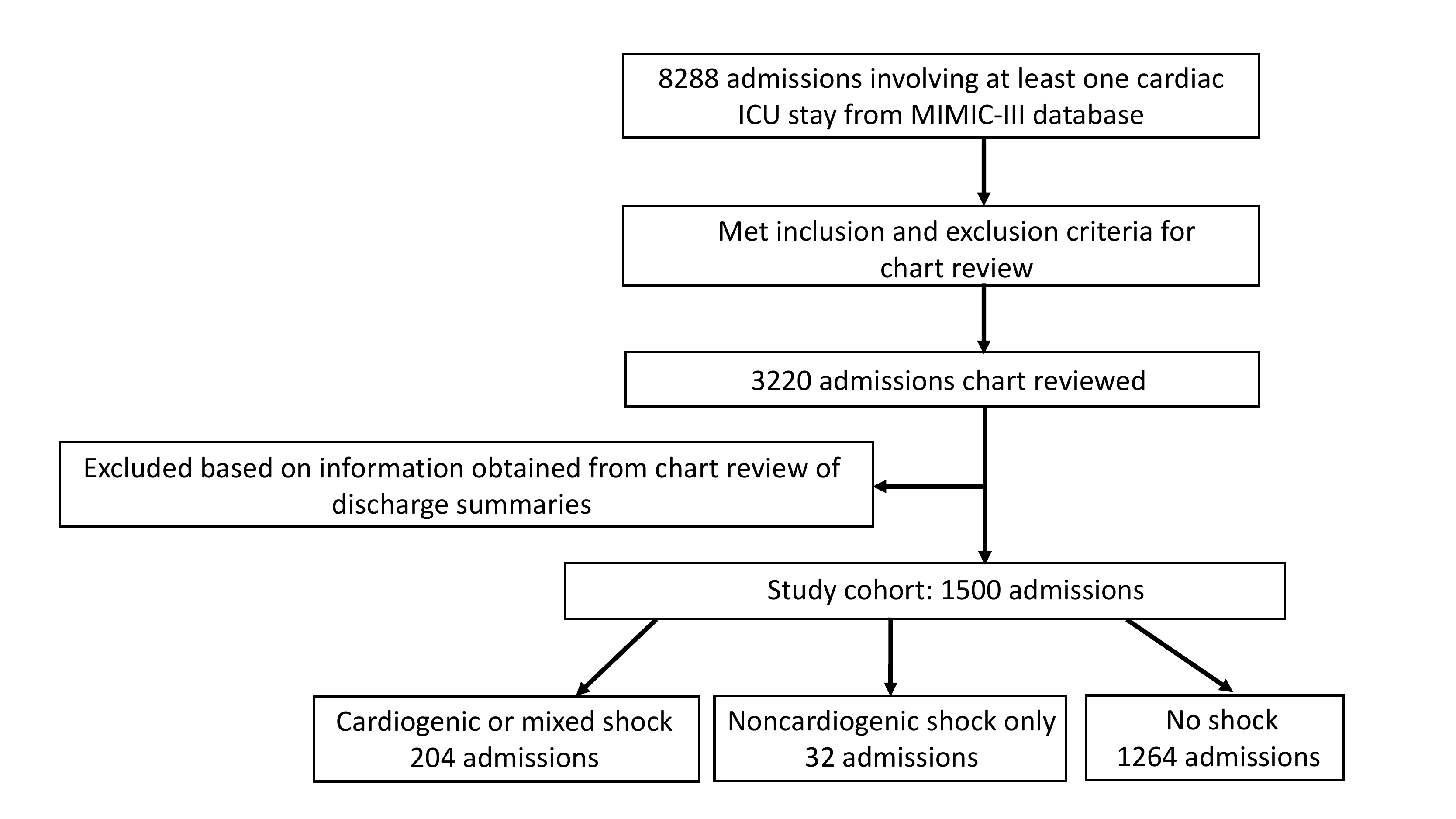}}
\caption{\label{fig:figure1}Study Cohort}
\end{figure}

We used a publicly available critical care database MIMIC-III \cite{johnson2016mimic}, which has 8188 admissions involving at least one cardiac \gls{icu} stay during the hospitalization (\cref{fig:figure1}). 3220 admissions met the inclusion and exclusion criteria for chart review (\cref{fig:figure1}): patients needed to have a possible diagnosis of acute decompensated heart failure and/or myocardial infarction with or without ST-elevation to be included; patients with age less than 18 years old or more than 89 years old, total hospital stay less than 24 hours, surgery \gls{icu} admission prior to cardiac \gls{icu} admission during the same hospitalization, shock on arrival to cardiac \gls{icu} were excluded (details in Methods, section Chart review).  The discharge summary and echocardiogram reports of the 3220 admissions were subsequently split among four physicians to review, which provided information regarding additional exclusion criteria, admission diagnoses, outcomes (no shock versus noncardiogenic shock only versus cardiogenic shock/mixed shock), left heart catheterization, and echocardiogram data (details in Methods, section Chart review). After applying the additional exclusion criteria (details in Methods, section Study cohort) obtained from chart review, the final study cohort comprised 1500 patients (\cref{fig:figure1}). 1264 had no shock event before cardiac \gls{icu} discharge; 204 had cardiogenic/mixed shock; 32 had noncardiogenic shock only (\cref{fig:figure1}). Patients' outcomes were retrospectively adjudicated from chart review. Time of shock onset was determined as the earliest time after cardiac \gls{icu} admission when 1) systolic blood pressure was less than 90mmHg for at least 30 minutes and there was evidence of systemic hypoperfusion or 2) pharmacologic agents/mechanical circulatory support was initiated to maintain systolic blood pressure above 90mmHg (\Cref{tab:hypoperfusion}) \cite{reyentovich2016management}. If there was discrepancy between physician adjudicated outcomes and objective data, i.e. physician adjudicated outcome revealed no shock but there was shock based on structured data, a second physician reviewed both the discharge summary and the structured data to reconcile the discrepancy (details in Methods, section Chart review).
\subsection*{Development of a dynamic risk score for prediction of cardiogenic shock} 
We developed CShock to determine the risk a patient has for developing cardiogenic shock after admission to the cardiac \gls{icu}. CShock is updated every hour after cardiac \gls{icu} admission. The features used to train the model for CShock included patient demographics, cardiac \gls{icu} admission diagnoses, routinely measured laboratory values and vital signs, and relevant features manually extracted from echocardiogram and left heart catheterization reports (Supplementary Information, List of features (\cref{suppsec:features})). \\

The model that underlies CShock was based on a dilated causal \gls{cnn} architecture, often used for time series modeling (See Methods, Model development for more details on model architecture). There were 194 input variables, including 182 time-varying (physiological) and 12 static (e.g., demographic) features. The physiological time series of the 182 time-varying features from cardiac \gls{icu} admission to event (discharge if no shock; shock onset otherwise) were inputted into the CShock model. The time-dependent variables also had missing indicators that took a value of 1 if the corresponding feature value was missing at each unique hour after cardiac \gls{icu} admission; 0 otherwise.\\

We designed a novel loss function specifically for risk scoring shown in \cref{eqn:main_loss} of the Supplementary Information. At a high level, optimizing the loss increases the soft-maximum risk across time for patients who developed cardiogenic shock during their stay while minimizing for those who did not develop cardiogenic shock. The study cohort (1500 patients) were split 50\%, 25\%, and 25\% into training, validation, and testing in a 4-fold cross validation. To improve performance, the model was pretrained in an auxiliary task of predicting in-hospital mortality for \gls{icu} patients in MIMIC-III database.\\

\begin{figure}[!ht]
    \centering
{\includegraphics[width=\textwidth]{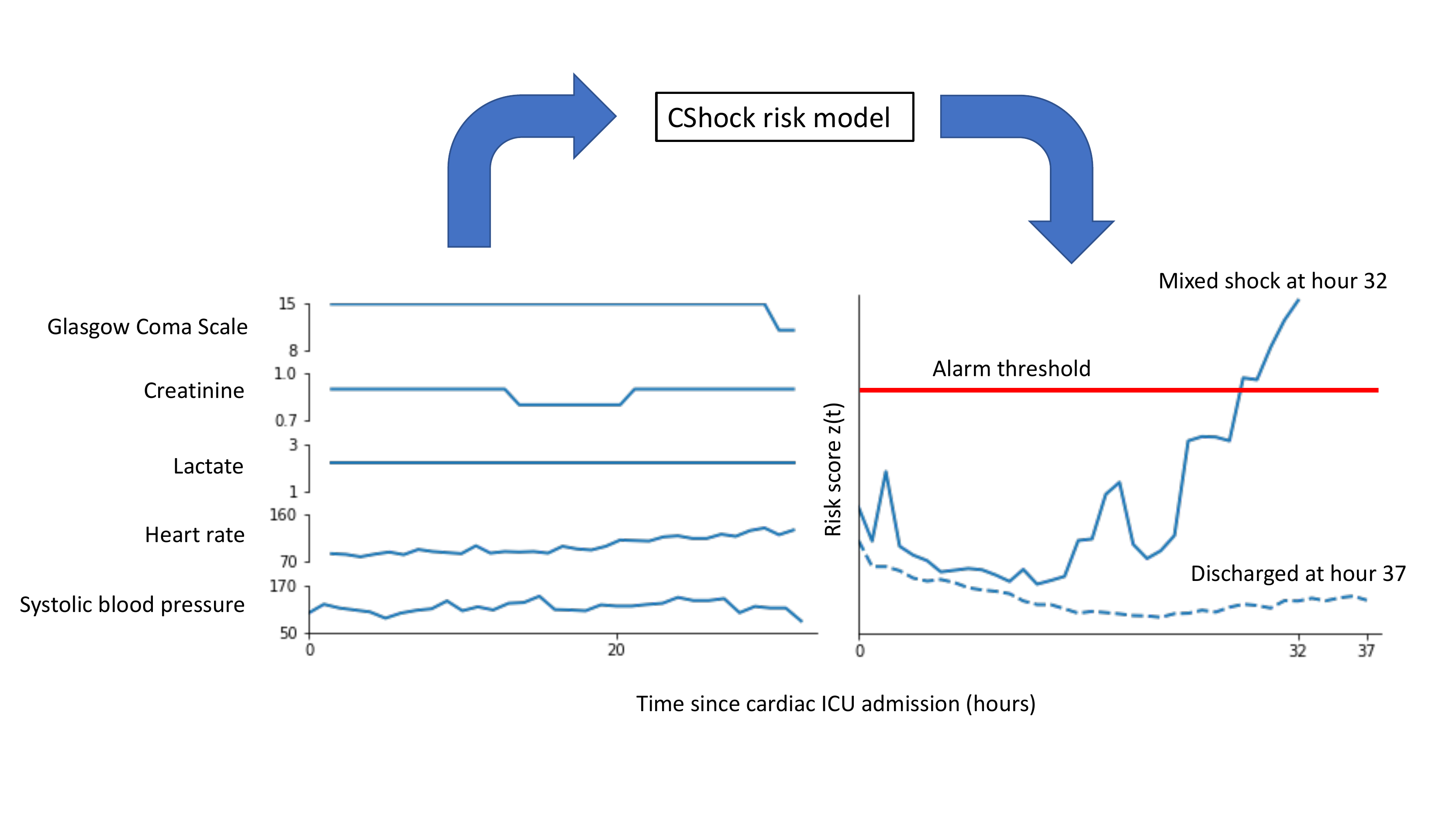}}
\caption{\label{fig:figure2} The physiological time series of five example features from cardiac \gls{icu} admission to event (discharge if no shock; shock onset otherwise) are shown for a patient developing mixed cardiogenic/noncardiogenic shock (left panel). Features displayed are Glasgow Coma Scale (a score for coma severity, see \cref{suppsec:features} for the three
components of the score), creatinine, lactate, heart rate and systolic blood pressure without support. The physiological time series were fed into the \gls{cnn}-based model, which outputted CShock scores. CShock scores (blue lines in right panel) were calculated for each hour from cardiac \gls{icu} admission until event. Two patients' CShock scores are shown here: one patient develops no shock and gets discharged from cardiac \gls{icu} at hour 37 (dashed blue line); the other patient goes into mixed shock at hour 32 (solid blue line). Cardiogenic shock detection occurs if a patient's CShock score exceeds the alarm threshold value. The horizontal red line indicates the detection threshold corresponding to a sensitivity of 0.8.}
\end{figure}

The left panel of \cref{fig:figure2} demonstrates the physiological time series of five example features that were entered into the dilated \gls{cnn}-based model, which then outputted the CShock score as shown by the blue lines in the right panel. The dashed blue line is the CShock score for a patient who develops no shock and gets discharged from the cardiac \gls{icu} at hour 37; the solid blue line shows the CShock score for a patient who goes into mixed cardiogenic/noncardiogenic shock at hour 32. Cardiogenic shock detection occurs if a patient’s CShock score exceeds the alarm threshold value. We followed MINIMAR (MINimum Information for Medical AI Reporting) \cite{hernandez2020minimar} in reporting the results.

\subsection*{Cardiogenic shock early warning system} 
\begin{figure}[!ht]
    \centering
{\includegraphics[width=0.7\textwidth]{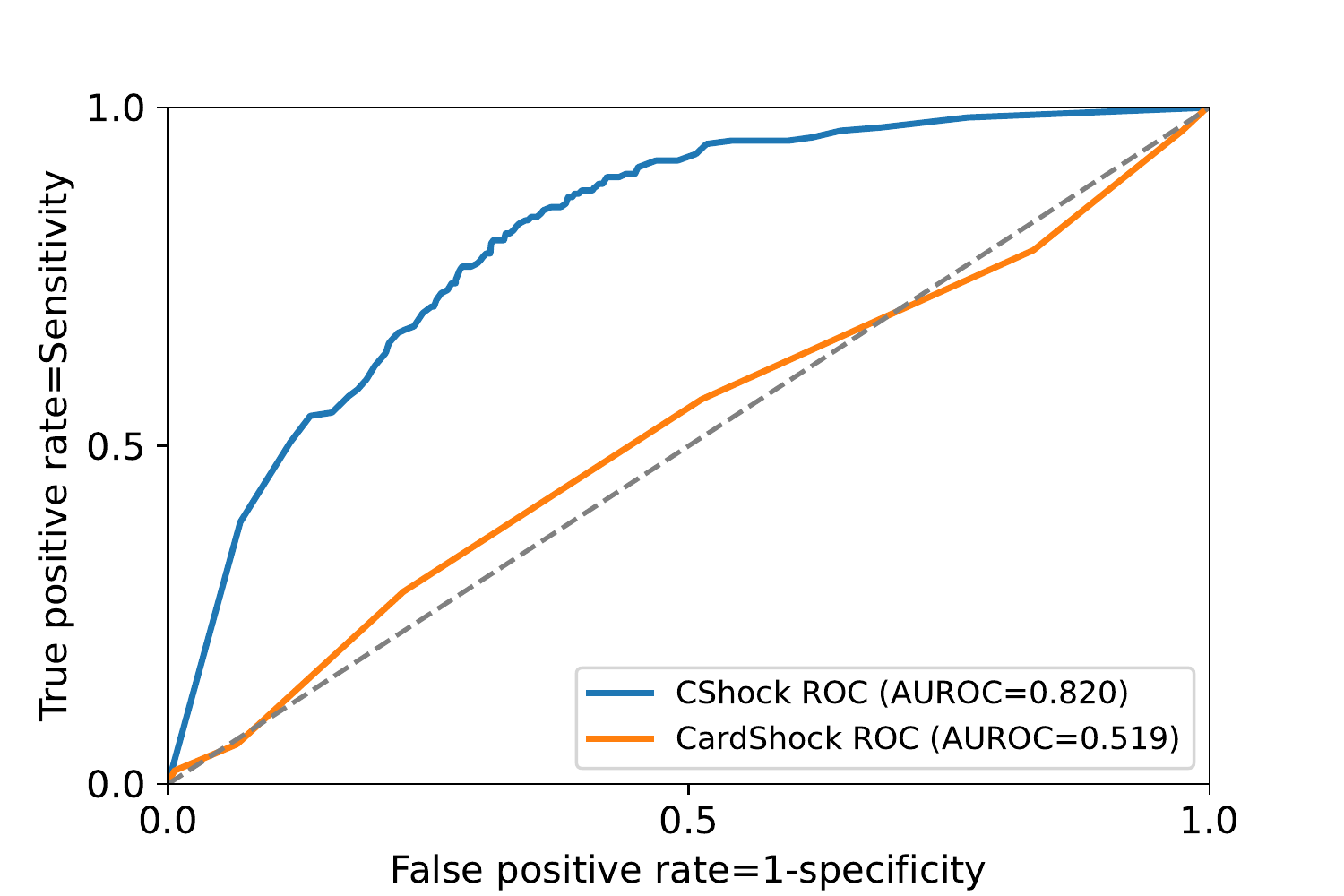}}
\caption{
\label{fig:figure3}\gls{roc} for predicting cardiogenic shock using the developed CShock score (blue) versus CardShock Score (orange) }
\end{figure}

We developed the CShock score for identifying patients at risk for cardiogenic shock before its overt onset. A patient was considered positive if the patient developed cardiogenic or mixed shock during their cardiac \gls{icu} stay; 204 patients were considered positive (\cref{fig:figure1}). The 1296 patients who developed noncardiogenic shock only or no shock in the study cohort were treated as negative cases.  The CShock score was able to achieve an \gls{auroc} of 0.820 (\cref{fig:figure3}). The \gls{roc} curve and its corresponding \gls{auc} were obtained by varying the alarm threshold that determined which patients were identified by the model as at risk for cardiogenic shock. At a positive predictive value of 0.33, CShock achieved a recall of 0.67 and was able to predict cardiogenic onset, on average, 36.5 hours before onset.\\

Current cardiogenic shock scores are mostly for outcome prediction once a diagnosis of cardiogenic shock has been made \cite{kalra2021risk}. We decided to use the CardShock score \cite{harjola2015clinical} to compare its performance to  CShock. Even though the CardShock score was not calibrated for our study objective, it is a widely accepted score for cardiogenic shock and was derived from a cohort of both \gls{acs} and non-\gls{acs} etiology, similar to the population in our study. The \gls{auroc} for predicting development of cardiogenic shock using the CardShock score was 0.519 (\cref{fig:figure3}). The basis for the poor performance of the CardShock score for this purpose, especially when compared to that for in-hospital mortality \cite{harjola2015clinical}, could be multifold: 1) history of myocardial infarction (a variable in the score calculation) is not available for the study cohort and thus the CardShock score was underestimated for patients with a history of myocardial infarction 2) the score is meant for prognosis once the diagnosis of cardiogenic shock has been made and not for risk stratifying patients for development of cardiogenic shock, i.e. variables such as low glomerular filtration rate are associated with worse mortality outcome but are not necessarily the most predictive for development of cardiogenic shock.

\subsection*{Model performance in different subgroups}
We performed subgroup analysis based on age and gender (\Cref{tab:rocauc}). The model performed better for the younger group than the older group. This could be due to the fact that the elderly tend to have more comorbidities that were embedded only in the unstructured data of the \gls{ehr}  (e.g. clinical notes) and thus unaccounted for in the model. We also found that the model performed better for male patients (0.817) compared to female patients (0.756). This was possibly a result of the study cohort having more male patients (918) as compared to female patients (582) and thus the model was better trained for the male cohort.

\begin{table}[!tb]
\centering
\begin{tabular}{llll}
\toprule
&  &\gls{auroc} \\
\midrule
\textbf{Age (years)} & below 25 percentile (less than 59) & 0.855$\pm$0.028\\
& between 25-50 percentile (at least 59, less than 70) & 0.801$\pm$0.060 \\
& between 50-75 percentile (at least 70, less than 80) & 0.775$\pm$0.035 \\
& above 75 percentile (at least 80) & 0.745$\pm$0.039 \\ 
\midrule
\textbf{Gender} & Female & 0.756$\pm$0.038  \\
&Male & 0.817$\pm$0.026  \\
\bottomrule
\end{tabular}
\caption{%
\textit{\gls{auroc} for each age group and gender}
\label{tab:rocauc}
}
\end{table}

\subsection*{External validation} 
The model was externally validated using a dataset that comprised all patients admitted to the cardiac \gls{icu} at \gls{nyu} Langone Health who met the same inclusion and exclusion criteria as the study cohort (\cref{fig:figureS1}). Patients' outcomes in the external validation cohort were retrospectively adjudicated by a physician. 58 of the 182 time-varying input features in the original model are routinely measured in the \gls{ehr} at \gls{nyu} Langone Health. Therefore, we developed the reduced model by training with 70 input features (58 were time-varying, 12 static) using MIMIC-III database (list of features is available in the Supplementary Information (\cref{suppsec:features}), denoted as included in the reduced model). The reduced model was able to achieve similar \gls{auroc} at 0.802 on the study cohort. It was then externally validated using the \gls{nyu} Langone Health cardiac \gls{icu} dataset and obtained an \gls{auroc} of 0.800, demonstrating its generalizability and applicability in other cardiac \glspl{icu}.

\subsection*{Model interpretation} 
\begin{figure}[t]
    \centering
    \begin{subfigure}[c]{0.55\textwidth}
        \vspace{1.0em} 
        \caption{\label{fig:fig4a}}
        \includegraphics[width=\textwidth]{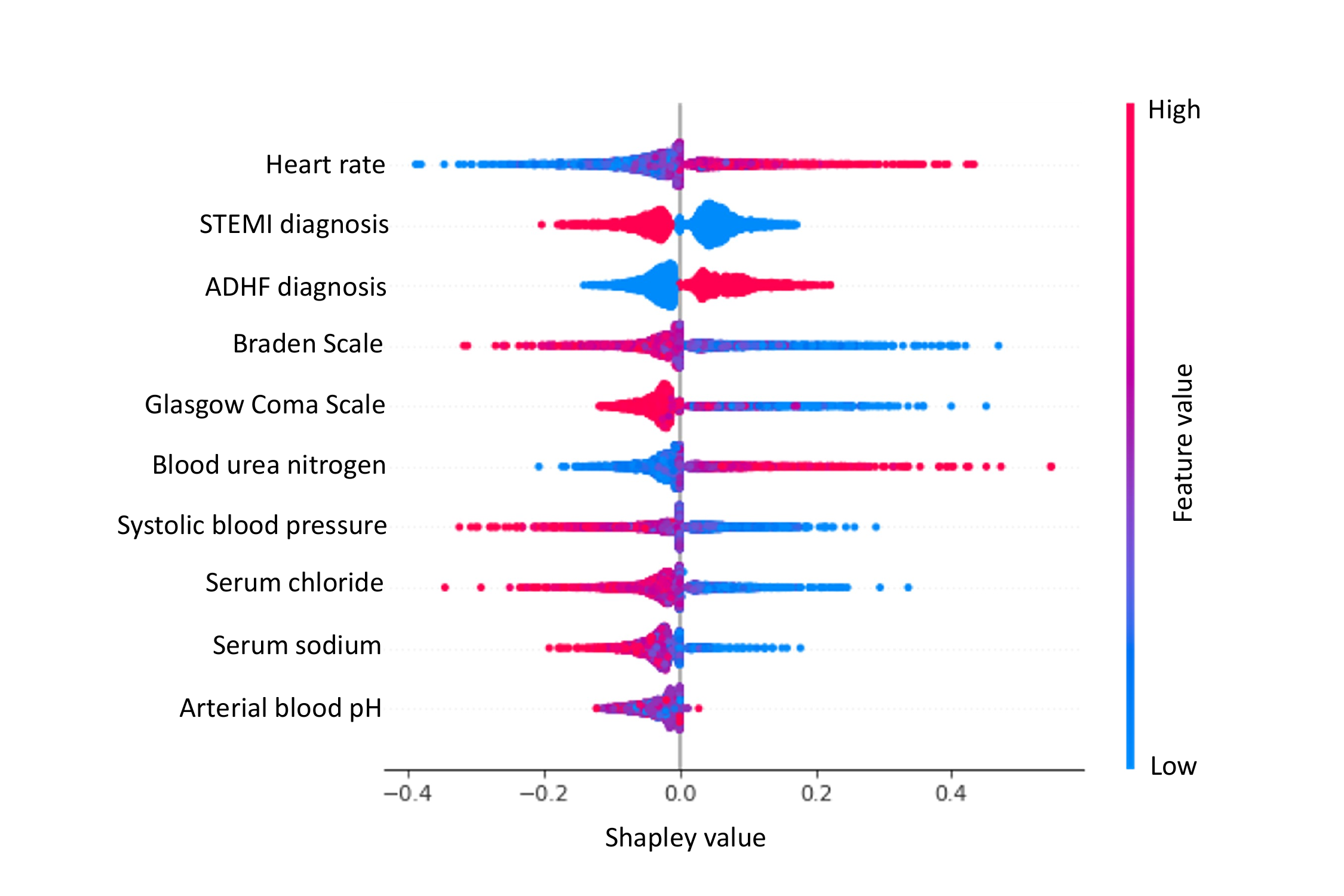}
    \end{subfigure}
    \hfill
    \begin{subfigure}[c]{0.44\textwidth}
        \caption{\label{fig:fig4b}}
        \includegraphics[width=\textwidth]{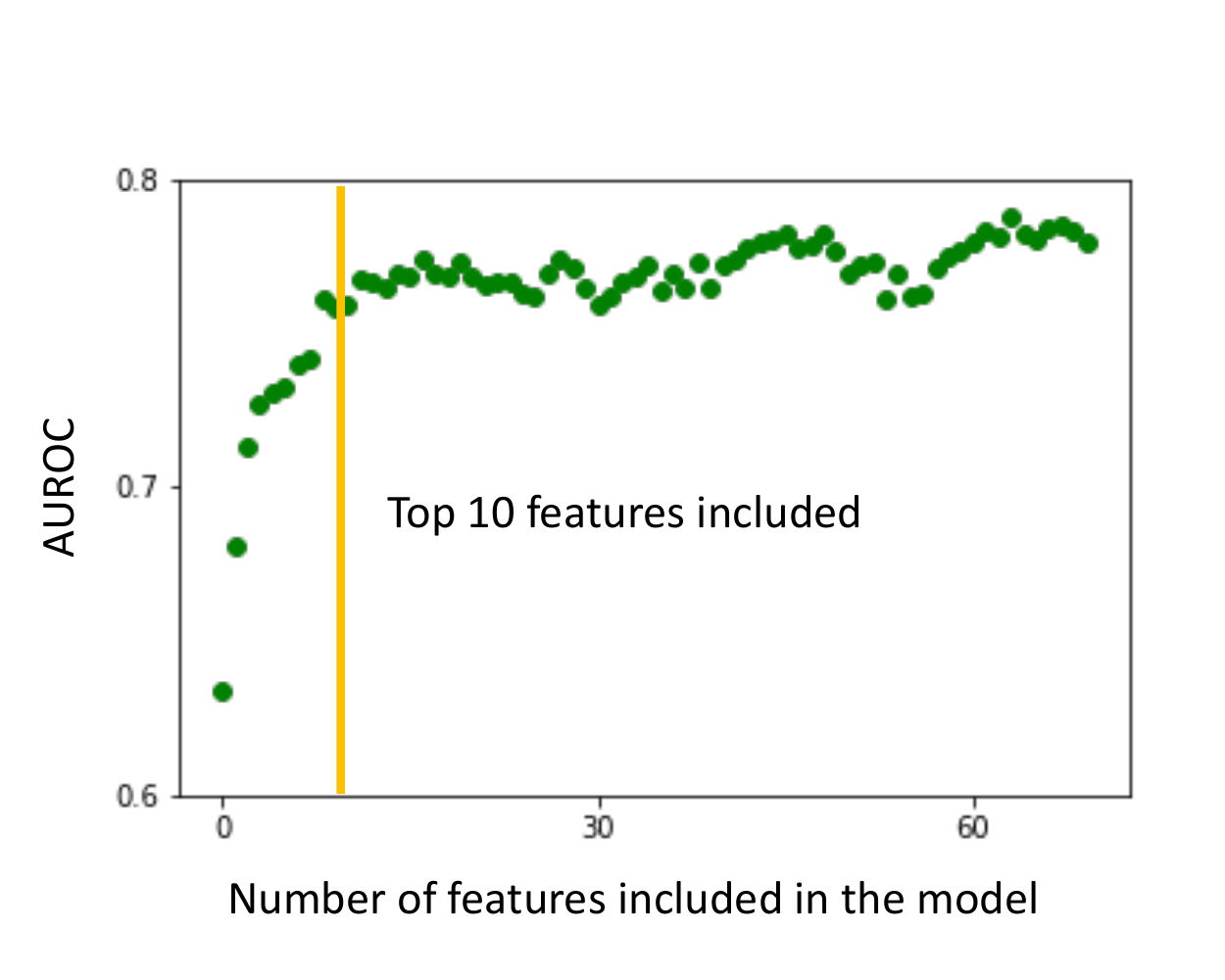}
    \end{subfigure}
    \caption{
        \cref{fig:fig4a} shows the 10 most important features based on Shapley values in descending order of importance (1 Heart rate, 2 having an admission diagnosis of myocardial infarction with ST-segment elevation, 3 having an admission diagnosis of acute decompensated heart failure, 4 Braden Scale, 5 Glasgow Coma Scale, 6 Blood urea nitrogen, 7 Systolic blood pressure, 8 Serum chloride, 9 Serum sodium, 10 Arterial blood pH). Each dot represents the feature values averaged over time of an individual patient from the training dataset with a higher value being more red and a lower value being more blue. Positive and negative Shapley values are associated with an increase or decrease in risk of cardiogenic shock development, respectively. \cref{fig:fig4b} \gls{auroc} obtained with increasing number of features included in the FastSHAP surrogate model using the evaluation dataset. Yellow line indicates the \gls{auroc} when only the 10 most important features (as displayed in \cref{fig:fig4a}) were included in the FastSHAP surrogate model that was trained to predict on subsets of features.
        \label{fig:fig4ab}
    }
\end{figure}

We used Shapley values \cite{lundberg2017unified} to provide explanations for how each feature influenced predictions. We employed FastSHAP \cite{jethani2021fastshap}, an efficient algorithm for calculating Shapley values \cite{lundberg2017unified}. The Shapley values were based on a FastSHAP's surrogate model that was trained to produce CShock risk scores given different feature subsets by masked prediction \cite{covert2021explaining,jethani2021have} rather than model the covariates \cite{sudarshan2020deep} (See \cref{suppsec:interpret} for more details on model interpretation). Shapley values were determined for the 70 features in the reduced model using the training dataset. \Cref{fig:fig4a} presents the ten most important features, i.e. with the largest Shapley values; \cref{fig:fig4b} shows how the \gls{auroc} changes as more features are added in descending order of importance into the FastSHAP surrogate model. \Gls{auroc} improved as more features were included in the model and the best \glspl{auroc} were similar to that obtained using the reduced model. With only the 10 most important features (as displayed in \cref{fig:fig4a}), FastSHAP surrogate model was able to achieve \gls{auroc} (yellow line in \cref{fig:fig4b}) similar to that when more features were included.

Returning to \cref{fig:fig4a}, each dot represents the feature values averaged over time of an individual patient from the training dataset with a higher value being more red and a lower value being more blue. Positive and negative Shapley values are associated with an increase or decrease in the CShock score, i.e. risk of cardiogenic shock development, respectively. For instance, heart rate is the most predictive feature of cardiogenic shock development and having an elevated heart rate is associated with an increased risk of cardiogenic shock development. This model interpretation analysis indicated that an admission diagnosis of myocardial infarction with ST-segment elevation is associated with lower risk of development of cardiogenic shock, whereas having an admission diagnosis of acute decompensated heart failure is associated with higher risk of development of cardiogenic shock. This likely reflects advancements in \gls{acs}  management strategies over the years, which makes \gls{acs} patients less likely to develop cardiogenic shock. Among the 10 most important features, low Braden Scale (a risk score for identifying patients at risk for pressure ulcers \cite{bergstrom1987braden}, see \cref{suppsec:features} for the six variables in Braden Scale), Glasgow Coma Scale, systolic blood pressure without support and serum sodium are predictive of cardiogenic shock development, as expected clinically \cite{kataja2018altered,leier1994clinical,naidu2022scai}. Lactate, a commonly used lab measurement to assess for severity of cardiogenic shock, is not among the 10 most important features; conceivably other top features captured most of the predictive information that would be embedded in lactate.\\

\section*{Discussion}
Myocardial infarction and heart failure affect 9 and 6 million people in the US respectively \cite{virani2021heart}. Cardiogenic shock is a common complication of myocardial infarction and heart failure and its occurrence is associated with substantial health and economic costs \cite{goldberg2009thirty,rossello2021synergistic}. Early recognition of cardiogenic shock is of paramount importance as it facilitates timely treatments that could potentially prevent the vicious spiral of cardiogenic shock and reduce the health and financial burden. However, early identification of cardiogenic shock has remained challenging \cite{papolos2021management}. We have demonstrated that the CShock score can predict cardiogenic shock with excellent \gls{auroc}. The early-warning system was tested in an independent patient cohort from a different hospital and showed comparable performance as in the development data.\\

Our CShock risk model is readily deployable. The reduced model with 70 features that are routinely captured in the \gls{ehr} database at \gls{nyu} Langone Health achieved an \gls{auroc} of 0.800. This model should be easily implemented at other facilities with modern \gls{ehr} systems. It is interesting to note that the reduced model performed on par with the full model despite having fewer features. This could be because 1) it is difficult to manually extract features from unstructured data (e.g. echocardiogram and catheterization data) and the lack of data can jeopardize the performance of the machine learning model; it is possible if those features were available as structured data, the performance of the full model would be even better. 2) The structured data captured most of the predictive information that would be embedded in unstructured data; incorporating unstructured data will not improve the performance of the machine learning model beyond what has been achieved with structured data only.\\

Previous machine learning studies relied on \gls{icd} codes and natural language processing of clinical notes \cite{henry2015targeted,liu2019data} to determine outcomes such as sepsis and septic shock, which are often inaccurate \cite{misset2008reliability}. In contrast, the outcome events for the cardiac \gls{icu} patients were carefully adjudicated by physicians in this study. This will allow the CShock score developed by this project to translate more easily into clinical implementation.\\

We constructed a novel loss function specifically for risk scoring. The dilated causal \gls{cnn} architecture along with the novel loss function are of broad applicability and can be employed for risk stratification in multiple clinical settings such as sepsis and pulmonary embolism. Physiological time series models have not widely employed pretraining, a common technique used to improve model performance in deep learning \cite{devlin2018bert,mcdermott2021comprehensive}.  Pretraining is useful when there is access to a large dataset labeled for a task related to---but not identical to--- the task of interest. In this case, we showed that pretraining with an auxiliary task of predicting mortality for \gls{icu} patients in MIMIC-III database improved the \gls{auroc} of the cardiogenic shock model from 0.752 to 0.820. We demonstrated that pretraining with a generic label such as mortality can be a useful strategy to improve physiological time series model performance when there is limited number of patients who meet all the inclusion and exclusion criteria of the study.\\

Shapley values are widely used to provide explanation for `blackbox'  models. However, they are computationally costly and are impractical to implement when large number of parameters are involved. We successfully implemented FastSHAP \cite{jethani2021fastshap}, a novel and efficient algorithm to estimate Shapley values to provide understanding about how each feature influenced cardiogenic shock prediction. Given the popularity of deep learning models in medicine and the `blackbox' nature of these models, we demonstrated an efficient yet effective way to uncover important facets of the deep learning models.\\

There have been other recent efforts to predict cardiogenic shock with machine learning \cite{chang2022early,rahman2022using}. However, they only considered patients who required inotropes/mechanical circulatory support as developing cardiogenic shock and used the time to initiate supportive measures as shock onset, which are not in alignment with the clinical criteria used in previous landmark trials \cite{hochman1999early,thiele2012intraaortic}; the early presentation of cardiogenic shock with low blood pressure and end-organ hypoperfusion would be missed by their algorithms. In addition, both studies lacked external validation to demonstrate generalizability in other populations. 1\% of the study cohort developed cardiogenic shock in the study by Rahman et al. \cite{rahman2022using}, which resulted in the best positive predictive value
being only 11\% (with recall/sensitivity 27\%) and would lead to the algorithm sounding many false alarms secondary to extreme class imbalance \cite{yang2020rethinking} and missing the majority of cardiogenic shock patients. The study by Chang et al. \cite{chang2022early} relied on \gls{icd} codes to determine outcomes, which would be inaccurate as discussed above; they excluded mixed cardiogenic/noncardiogenic shock patients from the study cohort, which could limit the algorithm's applicability.\\

Given the performance of our CShock score, we hypothesize that this \gls{cnn}-based early-warning model can help cardiac \gls{icu} teams clinically integrate complex data and more rapidly identify patients at risk for cardiogenic shock. A recent study using the Critical Care Cardiology Trials Network data showed that a multidisciplinary shock team approach improved outcomes in cardiogenic shock \cite{papolos2021management} and this machine learning based model could build upon this improvement, alerting shock teams to impending cardiogenic shock. Future prospective studies should be conducted to evaluate the generalizability of our findings across diverse health care systems and patient populations. In summary, we demonstrate that the CShock score we developed has the potential to provide automated detection and early warning for cardiogenic shock and improve the outcomes for the millions of patients who suffer from myocardial infarction and heart failure.

\section*{Acknowledgements}
YH was supported by a NIH T32 grant (T32HL098129). RR was supported by NIH/NHLBI Award R01HL148248, NSF Award 1922658 NRT-HDR: FUTURE
Foundations, Translation, and Responsibility for Data Science, NSF CAREER Award 2145542.
\bibliographystyle{unsrt}
\bibliography{shock}

\clearpage
\appendix
\section*{Methods}
\subsection*{Database and data extraction} 
We used a publicly available critical care database MIMIC-III, a single-center database that consists of patients admitted to critical care units at a large tertiary care hospital \cite{johnson2016mimic}. Data include vital signs, medications, laboratory measurements, notes by providers, and fluid balance. This database has been widely used to develop machine learning algorithms for risk stratifying critical care patients for sepsis and septic shock \cite{henry2015targeted,liu2019data}. \Gls{ehr} data was extracted from the MIMIC-III PostgreSQL database. The majority of data entries in MIMIC-III are comprised of timestamp-value pairs with a subject id, identifying which patient the data belong to, and an itemid identifying the meaning of the data.

\subsection*{Chart review} 
There are 8288 admissions in the MIMIC-III database that involve at least one cardiac \gls{icu} stay during the hospitalization (\cref{fig:figure1}). To be eligible for chart review, the admissions needed to have possible diagnosis of 1) acute decompensated heart failure and/or 2) myocardial infarction with or without ST-elevation. We used the following criteria to identify those admissions: 1) positive troponin or 2) \gls{icd-9} or 3) received intravenous diuretics such as furosemide or bumetanide or 4) \gls{icd-9} code for heart failure. The admissions that met the following criteria (based on structured data extracted from MIMIC-III database, \cref{fig:figure1}) were excluded from chart review: age less than 18 years old or more than 89 years old, total hospital stay less than 24 hours, surgery \gls{icu} admission prior to cardiac \gls{icu} admission during the same hospitalization, patients who presented in shock on arrival to cardiac \gls{icu}. 3220 admissions met the above inclusion and exclusion criteria and the discharge summaries and echocardiogram reports were obtained and split between four physicians to review.  Research electronic data capture (REDCap) database was set up for the chart review process.\\

Data obtained from chart review of discharge summaries provided the following information: whether patients met certain exclusion criteria as detailed below in section Study Cohort, cardiac \gls{icu} admission diagnoses 1) myocardial infarction with or without ST elevation and/or 2) acute decompensated heart failure, whether shock occurred during first cardiac \gls{icu} stay and if occurred, the etiology of the first shock (cardiogenic versus non cardiogenic versus mixed) and whether left heart catheterization was performed before the first cardiac \gls{icu} discharge and if performed, the locations of severe lesions, culprit lesion and any intervention undertaken.\\

Patients' outcomes (no shock versus noncardiogenic shock only versus cardiogenic shock/mixed shock) were adjudicated by a team of physicians from chart review of discharge summary. Time of shock onset was determined as the earliest time after cardiac \gls{icu} admission when the conditions in \Cref{tab:hypoperfusion} were satisfied \cite{reyentovich2016management}, as previously described in Results. If there was discrepancy between physician adjudicated outcomes and objective data, a second physician reviewed both the discharge summary and the structured data to reconcile the discrepancy. This could happen when there was a single low blood pressure reading and patient had acute kidney injury from recent contrast use in left heart catheterization. Based on structured data, the patient would be labeled as having a shock event but upon review of discharge summary, the patient had no shock event.\\

Data obtained from chart review of echocardiogram reports yielded information regarding left ventricular ejection fraction, right ventricular systolic function, right ventricular dilation, right ventricular systolic pressure, significant left sided valvular disease and presence of left ventricular aneurysm or pseudoaneurysm.

\subsection*{Study cohort} 
Based on information obtained from chart review of discharge summaries, the admissions that met the following criteria were excluded (\cref{fig:figure1}): development of acute respiratory distress syndrome, massive pulmonary embolism, cardiac tamponade and mechanical complications of myocardial infarction, urgent coronary-artery bypass surgery, cardiac surgery in the 7 days prior to first cardiac \gls{icu} admission, cardiac arrest in the 7 days prior to first cardiac \gls{icu} admission and admission to cardiac \gls{icu} for diagnosis other than myocardial infarction with or without ST-elevation or acute decompensated heart failure. Patients who developed shock within 4 hours of cardiac \gls{icu} admission were also excluded due to the assumption that the patient was likely in a peri-shock state at the time of admission.\\

Our final study cohort comprised 1500 admissions (\cref{fig:figure1}): 1264 patients did not develop shock; 204 patients developed cardiogenic or mixed shock and 32 patients developed noncardiogenic shock only.

\subsection*{Model development}

\paragraph{Features.} There were a total of 194 input features (182 were time-varying) that were either derived from routinely available measurements in the \gls{ehr}  or from chart review of discharge summaries and echocardiogram reports. The complete list of features is provided in the Supplementary Information (\cref{suppsec:features}). All features were standardized (zero mean, one standard deviation) before feeding into the model. The 182 time-varying variables had missing indicators, as described previously in Results. Missing values were imputed using population mean. 

\paragraph{Model architecture.} The model was based on dilated causal \gls{cnn}, which has been shown to be easier to train compared to \gls{rnn} in time series modeling \citemethods{oord2016wavenet}. The model architecture consisted of 4 dilated causal convolution layers with each followed by a rectified linear unit (ReLU) activation \citemethods{nair2010rectified}, batch normalization, and dropout. The output of the dilated causal \gls{cnn} captured the patient's overall physiological state at each hour. This output was followed by a fully connected layer and sigmoid function to produce the time-varying CShock risk score for cardiogenic shock.

\paragraph{Training.} We used the Adam \citemethods{kingma2014adam} optimizer with a learning rate of 1x$10^{-3}$ and batch size was set to 256. We evaluated loss (see \cref{eqn:main_loss} in the Supplementary Information) on the validation set for each epoch. Training was terminated after 50 epochs, which was shown to be sufficient for loss to plateau. The model was implemented using Pytorch. Hyperparameter optimization and model selection were conducted using 4-fold cross validation on the study cohort (1500 patients were split 50\%, 25\%, and 25\% into training, validation, and testing). The training set was balanced for the minority class. To improve the performance of the model, we pretrained the model for an auxiliary task of predicting in-hospital mortality for \gls{icu} patients. The pretraining cohort included all patients admitted to cardiac and medical \gls{icu} in MIMIC-III database, excluding the study cohort used in developing CShock risk model. Other exclusion criteria for pretraining cohort included patients with age less than 18 years old or more than 89 years old, total hospital stay less than 24 hours, and surgery \gls{icu} admission during the same hospitalization. We used the pretrained weights to initialize the CShock model parameters. In addition, we tried several forms of data augmentation and contrastive learning, but we did not include them in the final model as their inclusion gave similar performance at the expense of longer run time.

\paragraph{Model validation.} 
The best model was selected as the one achieving the best \gls{auroc} on validation set. Performance was then evaluated using the model on the testing set. The \gls{auroc} on test set across 4 folds was averaged and reported.

\subsection*{External validation} 
\small

\textbf{Compliance with ethical regulations}
\newline
We followed \gls{nyu} Langone Health human research protections policy, and
completed the \gls{nyu} self-certification form for research activities that may be classified as quality improvement. The work involving external validation met the \gls{nyu} Grossman School of Medicine \gls{irb} criteria for quality improvement work, not research involving human subjects, and thus this work did not require \gls{irb} review and no
informed consent was required or obtained.\\

\normalsize
The model was externally validated using a dataset that comprised all patients who met the same inclusion and exclusion criteria as described in section Study cohort admitted to the cardiac \gls{icu} at \gls{nyu} Langone Health. The external validation dataset consisted of 131 patients with 25 patients experiencing cardiogenic/mixed shock (\cref{fig:figureS1}). The average age of the external validation cohort was 64.5$\pm$14.9. There were 85 male and 46 female patients. The average age of the study cohort in MIMIC-III was 68.1$\pm$13.5. There were 918 male and 582 female patients in MIMIC-III study cohort. Race, ethnicity and socioeconomic status were not provided for the study cohort and the external validation cohort. 58 of the 182 time-varying input features in the original model are routinely measured in the \gls{ehr} at \gls{nyu} Langone Health and therefore, the model was retrained with 70 input features including 58 time-varying features using MIMIC-III database (a list of features in the reduced model is available in the Supplementary Information, (\cref{suppsec:features})). This model with reduced number of features was then externally validated using the dataset from the cardiac \gls{icu} at \gls{nyu} Langone Health. The admission diagnoses and outcomes (cardiogenic/mixed shock versus noncardiogenic shock only versus no shock) for these patients were obtained in the same manner as those for the study cohort.

\subsection*{Data availability}
The full MIMIC dataset can be downloaded from \url{http://physionet.org}. The labels for cardiogenic shock will be made available on request based on the MIMIC policies.

\subsection*{Code availability}
The computer code used in this research will be available at www.github.com/yuxuanhu12/cshock under an open-source license.

\subsection*{Author contribution}
Y.H., M.G., R.R. designed the experiments; Y.H., A.L. preprocessed and cleaned the MIMIC data; Y.H., A.T., C.T., S.M., J.M. reviewed the MIMIC charts; V.N., Y.A., N.K., J.Horowitz, T.A. selected and provided the \gls{nyu} clinical data and context; Y.H. preprocessed, cleaned, and chart reviewed the \gls{nyu} data; Y.H., M.G., R.R. developed the pipeline for CShock risk model; Y.H., M.G., implemented the CShock model; Y.H. implemented pretraining, analyzed subgroup performance and external validated the model with \gls{nyu} data; Y.H., M.G., N.J., R.R. devised and implemented the model interpretation strategy; M.S., A.P., N.S., S.K. contributed to various analyses of the data; Y.H., A.L., M.G., R.R. conceived and directed the project; Y.H., M.G. made the figures; Y.H., M.G., G.F., J.Hochman, S.K., S.B., R.R. wrote the manuscript.
\bibliographystylemethods{unsrt}
\bibliographymethods{shock}

\clearpage
\setcounter{figure}{0}
\renewcommand{\thefigure}{S\arabic{figure}}

\begin{figure}[!ht]
    \centering
{\includegraphics[width=\textwidth]{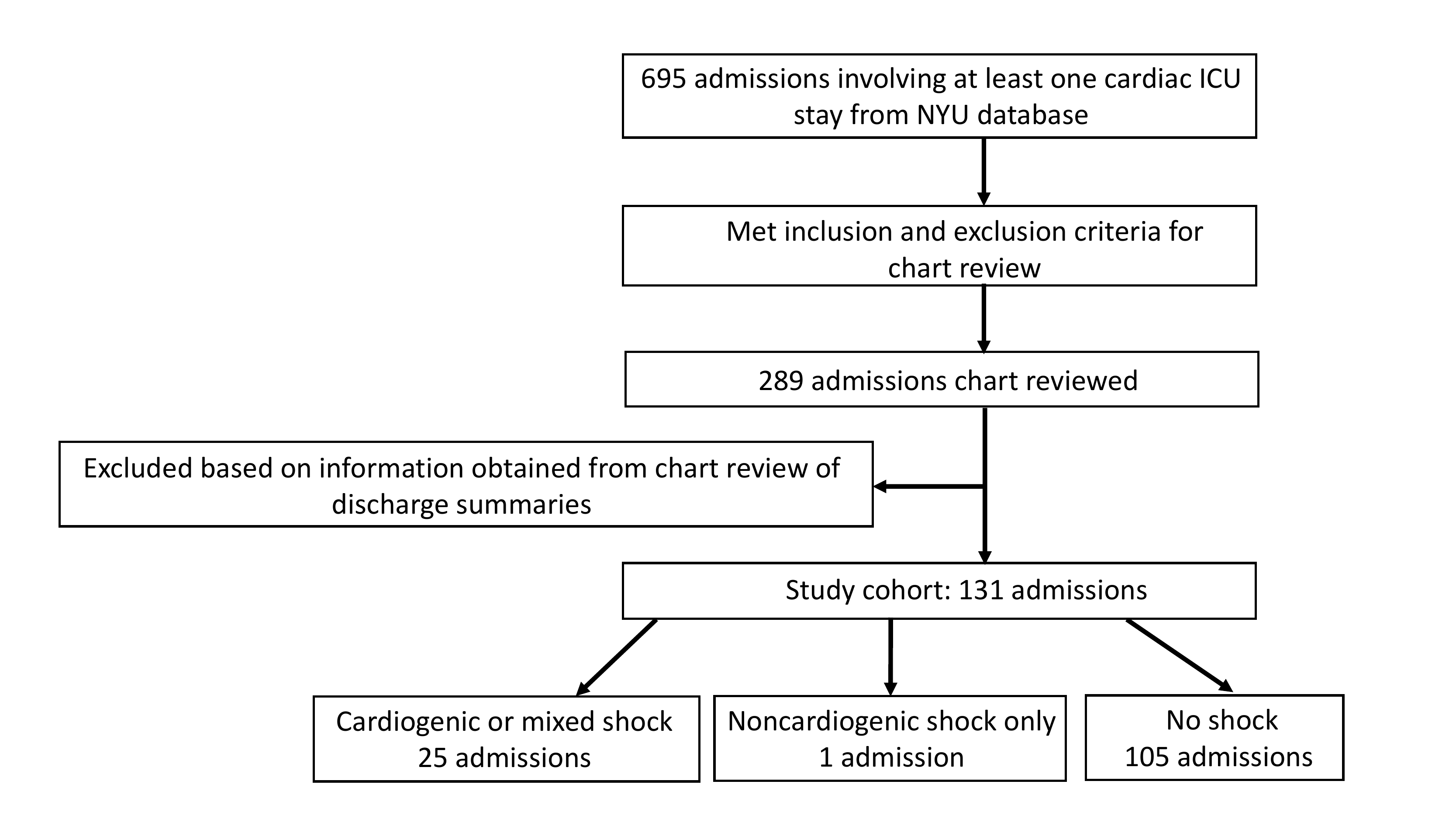}}
\caption{\label{figureS1}\gls{nyu} cardiac \gls{icu} Cohort for external validation}
\label{fig:figureS1} 
\end{figure}
\clearpage
\section{Supplementary Information}
\subsection{Loss function} 
CShock's loss function was constructed specifically for risk scoring. A risk score is calculated for each patient at each hour from time of admission until event of interest or discharge. For positive cases, the risk score is maximized for all times until the event of interest; for negative cases, the risk score is minimized until discharge. 

\paragraph{Notation.} We denote subject $i$'s observed event type as $E_i \in \{= \circ, \neq\circ\}$ where $E_i=\circ$ corresponds to cardiogenic/mixed shock and $E_i \neq \circ$ means the patient was not observed to develop cardiogenic shock, i.e. developed noncardiogenic shock only or no shock.
Each patient has observed time $U_i \in \mathbb{R}_{\geq 0}$, which could be the time of shock onset or discharge. At each time $k \leq U_i$, features $X_{i,\leq k}$ from time $0$ to time $k$ are observed. 

\paragraph{Model inputs.} The CShock risk model $r^\theta$ with parameters $\theta$ takes in a time $k$ and a variable-length feature list $X=[X_{i,0},\ldots, X_{i,k}]$ for $k \leq U_i$; it then produces CShock score for time $k$ denoted $r^{\theta}(X_{\leq k},k)$. Patient $i$'s modeled CShock scores for all times from admission up to $U_i$ 
are 

\begin{align}
    r_i^\theta[\leq U_i] &=
        \Big\{
        r_i^\theta(X_{0},0),
        \ldots,
        \underbrace{r_i^\theta(X_{\leq k},k)}_{\text{CShock score at time }k},
        \ldots,
        r_i^\theta(X_{ \leq U_i},U_i)
        \Big\}
\end{align}

\paragraph{Loss for patients with cardiogenic/mixed shock.}
For a patient with observed cardiogenic/mixed shock at time $U_i$, it is sufficient for \textit{any}  CShock score until $U_i$ to be large. We therefore maximize the maximum CShock score for patients with observed cardiogenic/mixed shock:
\begin{align}
    \max_\theta 
    \sum_i 
    \indicator{E_i = \circ}
    \Big( 
    \max_{k \in [0,\ldots,U_i]}
     r_i^\theta[\leq U_i]_k
    \Big) 
\end{align}
To make this maximization amenable to gradient-based optimization, we maximize the CShock scores at all times, rather than just the maximum-risk-so-far.
We weigh each risk by its proximity to the maximum risk and employ softmax for this purpose. For a patient with cardiogenic/mixed shock, the objective is:
\begin{align}
     \sum_k \text{Softmax}(r_i^\theta[\leq U_i])_k \cdot 
                        r_i^\theta[\leq U_i]_k
                        =
           \text{Softmax}(r_i^\theta[\leq U_i])^\top 
           r_i^\theta[\leq U_i]
\end{align}.

\paragraph{Loss for patients with noncardiogenic shock only or no shock.}
The loss for non-cardiogenic shock only/no shock patients is similar: we minimize the maximum CShock score for times $[0,\ldots,U_i]$ so that alarm does not sound for patients without cardiogenic shock.
The optimization problem is:
\begin{align}
    \min_\theta 
    \sum_i 
    \indicator{E_i \neq \circ}
    \Big( 
    \max_{k \in [0,\ldots,U_i]}
     r_i^\theta[\leq U_i]_k
    \Big) 
\end{align}
To relax this as a smooth optimization problem, we again use softmax weighting:
\begin{align}
     \sum_k \text{Softmax}(r_i^\theta[\leq U_i])_k \cdot 
                        r_i^\theta[\leq U_i]_k
                        =
           \text{Softmax}(r_i^\theta[\leq U_i])^\top 
           r_i^\theta[\leq U_i]
\end{align}

\paragraph{Combined loss function.}
Combining the loss for cardiogenic/mixed shock patients with that for noncardiogenic shock only/no shock patients, the loss, written as one minimization, is:
\begin{equation}
\label{eqn:main_loss}
\min_\theta\quad\sum_i \indicator{E_i\neq\circ}\text{Softmax}(\alpha r^{\theta}_{i}[\leq U_i])^\top r^\theta_{i}[\leq U_i]-\sum_i \indicator{E_i=\circ}\text{Softmax}(\alpha r^\theta_{i}[\leq U_i])^\top r^\theta_{i}[\leq U_i]
\end{equation}
where $\alpha$ (set to 2) is a temperature scaling constant in the softmax, which makes the softmax behave more closely to a true maximum selection.
Since cardiogenic shock detection occurs when CShock score exceeds the alarm threshold value, this loss function tightly maps to CShock's alarm performance.

\subsection{Model Interpretation: More Details \label{suppsec:interpret}} 
We used Shapley values, a popular explanation approach, to provide interpretations for model predictions.
A Shapley value quantifies on average how much \emph{a value function} changes 
when a feature 
is included versus when it is omitted in the presence of different subsets of the other features. 
Shapley values for the 70 features in the reduced model were estimated using FastSHAP \cite{jethani2021fastshap}.\\

Specifically, to construct a value function that scores different subsets of features, we train a surrogate model via masked prediction \citep{jethani2021have}. This surrogate model that takes as input a vector of masked feature pairs was trained by minimizing the same loss function as described in \cref{eqn:main_loss}. Let $\textbf{s}$ be a subset of features e.g. 
$\textbf{s}=\{1,8,11\}$. Let $\tilde{X}_t^{(\textbf{s})}$ denote the features 
$X_t$ at a given hour $t$, where only the features in $\textbf{s}$ are observed
and the rest are masked. Masking features means zeroing their value and setting their mask indicator to indicate masking (-1 in this case). Features were masked at a frequency of 0.5 in the training of the surrogate model. Then, we can define $r_i^\theta[\leq U_i](\textbf{s})$ to be equal to $r_i^\theta[\leq U_i]$ except
with $\tilde{X}^{\textbf{s}}$ instead of the original $X$. Finally, we can compute the original loss, but using $r_i^\theta[\leq U_i](\textbf{s})$ instead of $r_i^\theta[\leq U_i]$:
\begin{equation}
\label{eqn:surrogate_loss}
\min_\theta\quad\sum_i \indicator{E_i\neq\circ}\text{Softmax}(\alpha r^{\theta}_{i}[\leq U_i](\textbf{s}))^\top r^\theta_{i}[\leq U_i](\textbf{s})-\sum_i \indicator{E_i=\circ}\text{Softmax}(\alpha r^\theta_{i}[\leq U_i](\textbf{s}))^\top r^\theta_{i}[\leq U_i](\textbf{s})
\end{equation}
\newline
The value function we use in FastSHAP is the contribution of each feature to the loss function without the label: 
\begin{equation}
\label{eqn:value_fcn}
v(\textbf{s}) = \text{Softmax}(\alpha r^{\theta}_{i}[\leq U_i](\textbf{s}))^\top r^\theta_{i}[\leq U_i](\textbf{s})
\end{equation}
This value function tracks how the soft-maximum risk score changes as a function of different subsets of features $\textbf{s}$ for the $i$-th patient.
By using the trained surrogate model and the value function in \cref{eqn:value_fcn}, FastSHAP approximates Shapley values for the 70 features in the reduced model. The rankings for features based on Shapley values were computed by taking the mean absolute value across samples.

\subsection{Features used in the models \label{suppsec:features}} 
See data file attached

\end{document}